\documentclass{fcs}
\usepackage{bm}
\usepackage{stfloats}
\usepackage{subfig}
\usepackage{algorithmic}
\usepackage{algorithm}
\usepackage{booktabs}

\volumn{ }
\doi{ }
\articletype{REVIEW~ARTICLE}
\copynote{{\copyright} Higher Education Press and Springer-Verlag Berlin Heidelberg 2012}
\ratime{Received month dd, yyyy; accepted month dd, yyyy}
\email{xutianhao2018@gmail.com}
\title{Graph-Segmenter: Graph Transformer with Boundary-aware Attention for Semantic Segmentation}
\author{
Zizhang~Wu\xff$^{1}$,~
Yuanzhu~Gan$^{1}$,~
Tianhao~Xu$^{1,2}$,~
Fan~Wang$^{1}$}

\address{{$1$\quad Computer Vision Perception Department of ZongMu Technology, Shanghai, 201203, China.}\\
{$2$\quad Faculty of Electronics, Information Technology, Physics, Technical University of Braunschweig, \\Braunschweig, 38106, Germany.}}

\begin{document}
\maketitle
\setcounter{page}{1}
\setlength{\baselineskip}{14pt}

\begin{abstract}
The transformer-based semantic segmentation approaches, which divide the image into different regions by sliding windows and model the relation inside each window, have achieved outstanding success. However, since the relation modeling between windows was not the primary emphasis of previous work, it was not fully utilized. To address this issue, we propose a Graph-Segmenter, including a Graph Transformer and a Boundary-aware Attention module, which is an effective network for simultaneously modeling the more profound relation between windows in a global view and various pixels inside each window as a local one, and for substantial low-cost boundary adjustment. Specifically, we treat every window and pixel inside the window as nodes to construct graphs for both views and devise the Graph Transformer. The introduced boundary-aware attention module optimizes the edge information of the target objects by modeling the relationship between the pixel on the object's edge. Extensive experiments on three widely used semantic segmentation datasets (Cityscapes, ADE-20k and PASCAL Context) demonstrate that our proposed network, a Graph Transformer with Boundary-aware Attention, can achieve state-of-the-art segmentation performance.
\end{abstract}

\Keywords{Graph Transformer, Graph Relation Network, Boundary-aware, Attention, Semantic Segmentation}

\section{Introduction}
\noindent Semantic segmentation \cite{ruan2023intellectual_FCS_seg3,zhang2022multi_FCS_seg4} is a primary task in the field of computer vision, with the goal of labeling each picture pixel with a category that corresponds to it. As a result, intense research attention has lately been focused on it since it has the potential to improve a wide range of downstream applications \cite{grigorescu2020survey,feng2020deep,janai2020computer,arnold2019survey}, including geographic information systems, autonomous driving, medical picture diagnostics, and robotics. Modern semantic segmentation models almost follow the same paradigm\cite{wang2018understanding,devlin2018bert,wang2020dual,yu2020context}: they consist of a basic backbone for feature extraction and a head for pixel-level classification tasks in the current deep learning era. Improving the performance of the backbone network and the head of the segmentation model are two of the most controversial subjects in current semantic segmentation work.

Natural language processing (NLP) has been dominated by transformers for a long time \cite{rae2019compressive,lee2019set}, leading to a current spike of interest in exploring the prospect of using transformers in vision tasks, including significant advancements in semantic segmentation.
Using the vision transformer \cite{dosovitskiy2021an}, images are divided into a number of non-overlapping windows/patches, and some subsequent research investigated methods to enhance connections between windows/patches. It has the benefit of improving the modeling power of the system \cite{liu2021swin,xie2021segformer,chu2021twins}.

\begin{figure}[!t]
	\centering
	\includegraphics[width=8.4cm,height=5.6cm]{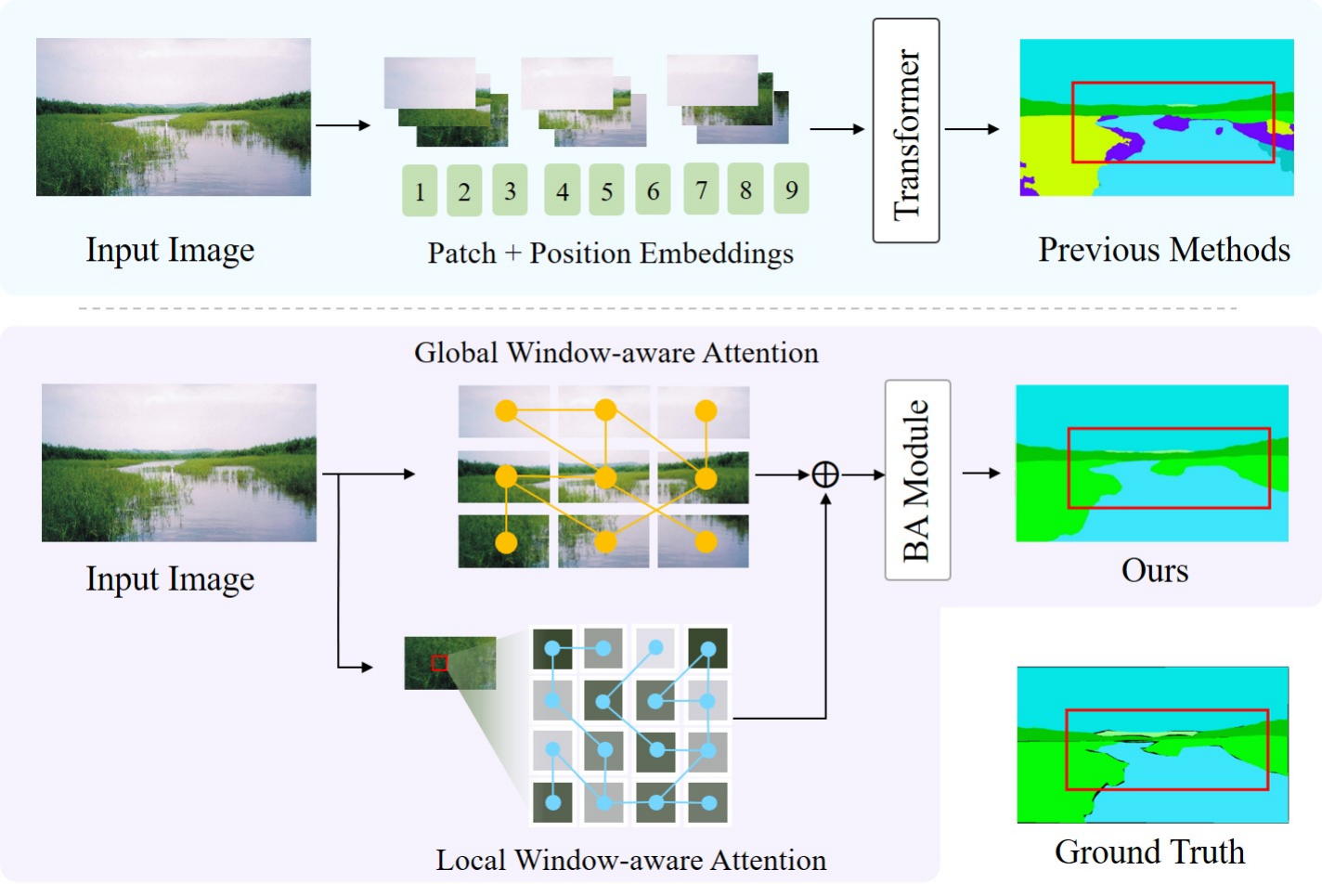}
	\vspace{-0in}
	\caption{An illustration of the proposed Graph-Segmenter with boundary-aware attention for semantic segmentation. 
	The top shows the segmentation results of the previous transformer-based semantic segmentation methods (e.g., Swin \cite{liu2021swin}). The bottom shows the actual segmentation results of our proposed Graph-Segmenter, which achieves promising boundary segmentation via the hierarchical level graph reasoning and efficient boundary adjustment requiring no additional annotation.}
	\label{fig:overview-diagram}
\end{figure}

Swin employs \cite{liu2021swin} a shifted window approach, however, the direction in which it interacts with other windows is fixed and unavoidably unresponsive.
Segformer utilizes \cite{xie2021segformer} the overlapping patch merging approach, however it focuses primarily on the local continuity between patches rather than the overall continuity between patches.
The modeling of long-distance interactions between windows/patches is not investigated in depth by these methodologies, which is unfortunate.
In addition to current work on the optimization of the backbone, a number of research have contributed to the design of the head in order to facilitate further optimization \cite{fang2021msg,wang2021kvt,chu2021we}. However, following the majority of them is computationally costly due to the fact that boundary prediction-based approaches need extra object boundary labeling for segmentation boundary optimization.

We propose a novel relation modeling method acting on sliding windows, using graph convolutions to establish relationships between windows and pixels inside each window, which enhanced the backbone to address the issues above. In particular, we regard each window or the pixels inside as nodes for the graph network and use the visual similarity between nodes to establish the edges between nodes. After that, we use the graph network further to update the nodes and edges of the graph. So that different nodes can adaptively establish connections and update information in network transmission to realize the nonlinear relationship modeling between different windows and different pixels inside. In brief, the network's overall feature learning and characterization capabilities are further improved by enhancing the long-distance nonlinear modeling capabilities between different windows and different pixels inside, as shown in Figure \ref{fig:overview-diagram}, which leads to an evident rise in performance.

Furthermore, we introduce an efficient boundary-aware attention-enhanced segmentation head that optimizes the boundary of objects in the semantic segmentation task, allowing us to reduce the labeling cost even further while simultaneously improving the accuracy of the semantic segmentation in the boundary of the objects under consideration. To put it another way, we develop a lightweight local information-aware attention module that allows for improved boundary segmentation. By determining the weights of the pixels around an object's border and applying various attention coefficients to distinct pixels via local perception, it is possible to reinforce the important pixels that are critical in categorization while weakening the interfering pixels. The attention module used in this study has just a few common CNN layers, which makes it efficient for segmentation boundary adjustments when considering the size, floating point operations, and latency time of the segmentation data.

We investigate the atypical interaction between windows that are arranged hierarchically on a vision transformer. Additionally, we improve the boundary of target instances using an efficient and lightweight boundary optimization approach that does not need any extra annotation information and can adaptively alter the segmentation boundary for a variety of objects. We have demonstrated in this paper that our Graph-Segmenter can achieve state-of-the-art performance on semantic segmentation tasks using extensive experiments on three standard semantic segmentation datasets (Cityscapes \cite{cityscapes}, ADE-20k \cite{ade20k}, and PASCAL Context \cite{PASCALC}), as demonstrated by extensive experiments on three standard semantic segmentation datasets.

\section{Related works}


\subsection{CNN-based Semantic Segmentation}
Convolutional neural networks (CNNs) based methods serve as the standard approaches throughout the semantic segmentation \cite{fcn,iotSeg4_SmartHealth,zhao2017pyramid,chen2018encoder,fu2019dual,ding2021interaction} task due to apparent advantages compared with traditional methods.
FCN \cite{fcn} started the era of end-to-end semantic segmentation, introducing dense prediction without any fully connected layer. Subsequently, many FCN-based methods \cite{zhao2017pyramid,chen2018encoder,yuan2020segmentation} have been proposed to promote image semantic segmentation. DNN-based methods usually need to expand the receptive field through the superposition of convolutional layers. Besides that, the receptive field issue was explored by several approaches, such as Pyramid Pooling Module (PPM) \cite{zhao2017pyramid} in PSPNet and Atrous Spatial Pyramid Pooling (ASPP) in DeepLabv3 \cite{chen2018encoder}, which expand the receptive field and capture multiple-range information to enhance the representation capabilities. Recent CNN-based methods place a premium on effectively aggregating the hierarchical features extracted from a pre-trained backbone encoder using specifically developed modules: DANet \cite{fu2019dual} applies a different form of the non-local network; SFNet \cite{li2020semantic} addresses the misalignment problem through semantic flow by using the Flow Alignment Module (FAM); CCNet \cite{huang2019ccnet} present a Criss-Cross network which gathers contextual information adaptively throughout the criss-cross path; GFFNet \cite{li2020gated}; APCNet \cite{he2019adaptive} proposes Adaptive Pyramid Context Network for semantic segmentation, which creates multi-scale contextual representations adaptively using several well-designed Adaptive Context Modules (ACMs). \cite{ding2021interaction} divides the feature map into different regions to extract regional features separately, and the bidirectional edges of the directed graph are used to represent the affinities between these regions, in order to help model region dependencies and alleviate unrealistic results. \cite{ding2019boundary} learns boundaries as an additional semantic category, so the network can be aware of the layout of the boundaries. It also proposes unidirectional acyclic graphs (UAGs) to simulate the function of undirected cyclic graphs (UCGs) to structure the image. Boundaryaware feature propagation (BFP) module can acquire and propagate boundary local features of regions to create strong and weak connections between regions in the UAG.

\subsection{Transformer and Self-Attention}
The attention mechanism for image classification was first seen in \cite{article14}, and then \cite{article15} employed similar attention on translation and alignment simultaneously in the machine translation task, which was the first application of the attention mechanism to the NLP field. Inspired by the dominant performance in the NLP field later, the application of self-attention and transformer architectures in computer vision are widely explored in recent years. Self-attention layers as a replacement for spatial convolutional layers achieved rises in terms of robustness and performance-cost trade-off \cite{article16}. Moreover, Vision Transformer (ViT) \cite{dosovitskiy2021an} shown transformer structure with simple non-overlap patches could surpass state-of-the-art results, and the impressive result led to a trend of vision transformers. The issue that ViT is unsuitable as a general backbone for semantic segmentation was further addressed by several approaches:
\cite{carion2020end} proposed DEtection TRansformer (DETR) for direct set prediction based on transformers and bipartite matching loss; based on DETR, \cite{zhu2021deformable} suggested Deformable DETR with attention modules that focus only on a limited number of critical sampling locations around a reference; after that, \cite{article20} investigated the video instance segmentation (VIS) task using vision transformers, termed VisTR, which view the VIS task as a direct end-to-end parallel sequence decoding/prediction problem.
\cite{detr3d} expands the idea of DETR to multi-camera 3D object detection to make great progress.

In addition, the transformer-based semantic segmentation methods \cite{strudel2021segmenter,zheng2021rethinking,liu2021swin,xie2021segformer} have drawn much attention. \cite{strudel2021segmenter} relies on a ViT \cite{dosovitskiy2021an} backbone and introduces a mask decoder inspired by DETR \cite{carion2020end}. \cite{zheng2021rethinking} reformulates the image semantic segmentation problem from a transformer-based learning perspective. \cite{liu2021swin} uses a shifted windowing scheme to break the limited self-attention computation into non-overlapping local windows. \cite{xie2021segformer} proposes a powerful semantic segmentation framework with lightweight multilayer perceptron decoders. They adopt the Overlapped Patch Merging module to preserve local continuity. However, these methods didn't look at the relationship between windows very well, which led this study to use hierarchical-level graph reasoning and efficient boundary adjustment. Our method puts more focus on the graph modeling between windows within the transformer blocks, which is the first work for the semantic segmentation task.

\subsection{Graph Model}
Graph-based convolutional networks (GCNs) \cite{zhang2019dual,pan2021weakly,wang2022local,wu2022graph} can exploit the interactions of non-Euclidean data, which was initially proposed in \cite{bruna2014spectral}. As graphs can be irregular, GCNs led to an increasing interest in recent years, and many excellent works further explored GCNs from various perspectives such as graph attention networks \cite{velivckovic2018graph} for guided learning, and dynamic graph \cite{zhang2020dynamic} to learn features more adaptively. 
GCNs also have achieved great performance in kinds of computer vision tasks, such as 
human pose estimation \cite{HumanPose}, image captioning \cite{imagecaptioning}, action recognition \cite{actionrecognition}, and so on. 
\cite{HumanPose} adopts graph atrous convolution and transformer layers to extract multi-scale context and long-range information for human pose estimation.
\cite{imagecaptioning} proposes an image captioning model based on a dual-GCN and transformer combination.
\cite{ImageCo-Saliency} proposes an adaptive graph convolutional network with attention graph clustering for image co-saliency detection. 
Moreover, GCNs can propagate information globally and model contextual information efficiently for semantic segmentation \cite{li2020spatial,Hu2020classwise,zhang2019dual,zhang2021affinity}, point cloud segmentation \cite{pointcloudsegmentation,su2022dla} and instance segmentation \cite{instancesegmentation}. 
\cite{li2020spatial} performs graph reasoning directly in the original feature space to model long-range context.
\cite{Hu2020classwise} introduces the class-wise dynamic graph convolution module to capture long-range contextual information adaptively.
\cite{zhang2019dual} captures the global context along the spatial and channel dimension by the graph-convolutional module, respectively.
\cite{zhang2021affinity} combines adjacency graphs and KSC-graphs by affinity nodes of multi-scale superpixels for better segmentation.
\cite{pointcloudsegmentation} proposes a hierarchical attentive pooling graph network for point cloud segmentation to enhance the local modeling.
\cite{review1,review2} provide more detailed surveys of GCNs.

\subsection{Discussion}
Different from Swin \cite{liu2021swin}, first of all, our proposed graph transformer focuses on the nonlinear relationship modeling of sliding windows, including the modeling of high-order (greater than or equal to 2-order) relationship between sliding windows and among different feature points inside each sliding window. Secondly, our graph transformer module is a lightweight pluggable sliding window modeling module, which can be embedded into some sliding window-based transfomer models such as Swin \cite{liu2021swin} to improve the feature modeling and expression ability of the original model. Finally, the graph transformer modules plus the boundary-aware attention modules together form our Graph-Segmenter network, which achieves state-of-the-art performance compared to other recent works on the existing segmentation benchmarks.

\begin{figure*}[htb]
	\centering
	\includegraphics[width=15.2cm, height=5.8cm]{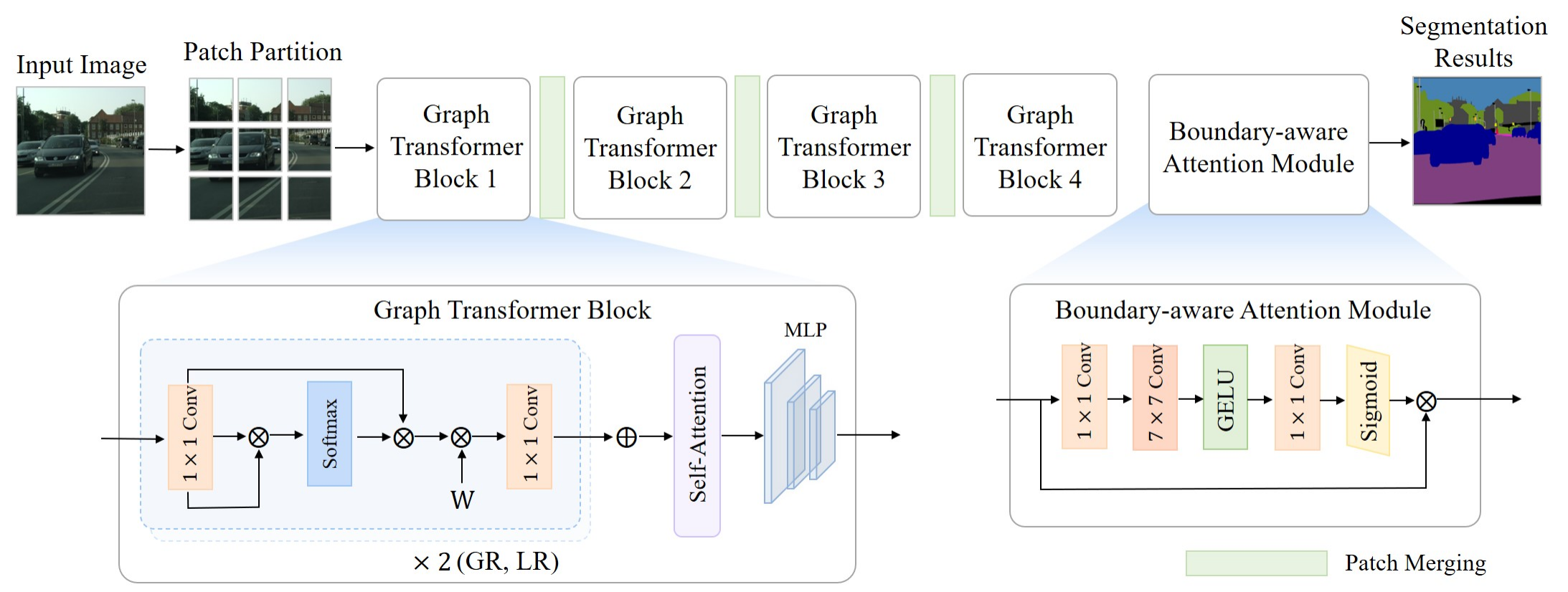}
	\vspace{-0in}
	\caption{An overview of the proposed Graph-Segmenter with efficient boundary adjustment for semantic segmentation, which includes Global Relation Modeling, Local Relation Modeling, and Boundary-aware Attention. ``GR'' denotes the global window-aware relation module and ``LR'' denotes the local window-aware relation module. ``GR'' or ``LR'' consists of $1\times 1$ convolution, softmax, $\bm{\mathrm{W}}$ and $1\times 1$ convolution. $\bm{\mathrm{W}}$ is the learnable weighting matrix, corresponding to $\bm{\mathrm{W}}^{(l)}$ in Equation \ref{gr_4}, and also equivalent to $\mathcal{C}^{gr}$ in ``GR'' or $\mathcal{C}^{lr}$ in ``LR''.}
	\label{fig:cib_arch}
\end{figure*}

\section{Methods}
\noindent This section will elaborate on the design of our devised Graph-Segmenter, which includes the Graph Transformer module and Boundary-aware attention module.

\subsection{The Overview}

As shown in Figure \ref{fig:cib_arch}, our proposed coarse-to-fine-grained boundary modeling network, Graph-Segmenter, includes a coarse-grained graph transformer module in the backbone and the fine-grained boundary-aware attention module in the head of the semantic segmentation framework. Assume that the input image passes through some CNN layers and then obtains the positional embedding enhanced feature maps $\bm{\mathrm{X}} \in \mathbb{R}^{C\times H\times W}$, where $C$ denotes the number of channels, $H$ and $W$ denote height and width in the spatial dimension of the image, respectively. In order to make better use of the Transformer mechanism, inspired by \cite{liu2021swin,chu2021twins}, we first divide feature maps $\bm{\mathrm{X}}$ into $M\times N$ window patches and define each window as $\{\bm{\mathrm{x}}_{m,n}\in \mathbb{R}^{C\times \frac{H}{M}\times \frac{W}{N}}|m\in\{1,2,...,M\},n\in\{1,2,...,N\}\}$. And then $\bm{\mathrm{X}}$ passes through Graph Transformer (GT) Network module, generating the relation-enhanced features $\bm{\mathrm{Y}}$ after GT module in the backbone network. After this, we use the boundary-aware attention (BA) module to adjust the segmentation boundary on the feature maps $\bm{\mathrm{Y}}$ and obtain the object boundary optimal feature $\bm{\mathrm{Z}}$. At last, we use these features $\bm{\mathrm{Z}}$ to formulate the corresponding segmentation loss function in each pixel. In the following subsections, we will elaborate on our proposed window-aware relation transformer network module and boundary-aware attention module. To facilitate the following description, we first introduce the efficient graph relation network, which plays a central role in the graph transformer block and is capable of efficiently and effectively modeling the nonlinear relations for abstract graph nodes.

\subsection{Efficient Graph Relation Network}
In this section, we first introduce the preliminary work on graph convolution networks (GCN), which introduces convolutional parameters that can be optimized compared to graph neural networks (GNN). As shown in \cite{velivckovic2018graph}, a GCN can be defined as
\begin{equation}\label{gr_00}
\bm{\mathrm{X}}=\bm{\mathrm{R}}\bm{\mathrm{X}}\bm{\mathrm{W}},
\end{equation}
where $\bm{\mathrm{X}}$, $\bm{\mathrm{R}}$ and $\bm{\mathrm{W}}$ are the input node matrix (here we still denote the output of the above equation as $\bm{\mathrm{X}}$), the adjacency matrix reflecting the relationship between nodes and the learnable convolution parameter matrix respectively.

Based on above preliminary work, we introduce an efficient graph relation network, which was inspired by \cite{velivckovic2018graph,hamilton2017inductive}, to model the non-linear relation of higher-order among the sliding windows and inside each window. Given the input $\bm{\mathrm{X}}$, we define the relation function $\mathcal{R}$ as follows
\begin{equation}\label{gr_0}
\bm{\mathrm{R}}=\mathcal{R}(\bm{\mathrm{X}}).
\end{equation}

As shown in Equation ($\ref{gr_0}$), to establish the relationship $\mathcal{R}$ between different input data $\bm{\mathrm{x}}_{i}$ and $\bm{\mathrm{x}}_{j}$, where $\bm{\mathrm{x}}_{i}, \bm{\mathrm{x}}_{j}\in \bm{\mathrm{X}}$, $i,j\in \{1,2, \dots, K\}$, and $K$ is the number of nodes in $\bm{\mathrm{X}}$, mathematically, we can establish a simple linear relationship using linear functions or a more complex nonlinear relationship using nonlinear functions. In general, linear relations are poor in terms of robustness, so here we design a quadratic multiplication operation that establishes a nonlinear relationship between different input nodes.

\begin{equation}\label{gr_1}
r_{i,j}=\frac{\bm{\mathrm{x}}_{i}\cdot \bm{\mathrm{x}}_{j}}{||\bm{\mathrm{x}}_{i}\cdot \bm{\mathrm{x}}_{j}||},
\end{equation}

where $r_{i,j} \in \bm{\mathrm{R}}$ is the learned relation between $\bm{\mathrm{x}}_{i}$ and $\bm{\mathrm{x}}_{j}$. $||\cdot||$ represents a two-norm operation. Because $\bm{\mathrm{x}}_{i}\cdot \bm{\mathrm{x}}_{j} \leq ||\bm{\mathrm{x}}_{i}\cdot \bm{\mathrm{x}}_{j}||$, $| r_{i,j} | \leq 1$, where $|\cdot|$ denotes the absolute value operation, we do not need to normalize the $r_{i,j}$ as in \cite{kipf2017semi}.

With the connection relation matrix $\bm{\mathrm{R}}$, we can define the node update function $\mathcal{U}$ of the graph as follows

\begin{equation}\label{gr_2}
\bm{\mathrm{X}}^{(l+1)}=\mathcal{U}(\bm{\mathrm{X}}^{(l)}).
\end{equation}

Now that we have $r_{i,j}$ and also to design an efficient graph relational network, we want this graph network node propagation update operation to be a sparse matrix operation. To achieve efficient computation, we use the indicative function and the relational matrix $\bm{\mathrm{R}}$. Equation (\ref{gr_2}) can be rewritten into a form that satisfies the sparse matrix operation as follows:
\begin{equation}\label{gr_3}
\bm{\mathrm{x}}_{i}^{(l+1)}=\sum\limits_{\bm{\mathrm{x}}_{j}^{(l)}\in \delta(\bm{\mathrm{x}}_{i}^{(l)})}\mathbb{I}(r_{i,j}>\theta)\cdot \bm{\mathrm{x}}_{j}^{(l)}, \\\\i,j\in \{1,2, \dots, K\},
\end{equation}
where $\mathbb{I}(\cdot)$ is the indicative function which takes $r_{ij}$ if the condition holds, or 0 otherwise. $\bm{\mathrm{x}}_{j}^{(l)}$ is the $j$th graph node in layer $l$ in the graph, which is generated in the layer-by-layer transfer process of graph convolutional network. ${\delta(\bm{\mathrm{x}}_{i}^{(l)})}$ denotes the set of neighboring nodes of node $\bm{\mathrm{x}}_{i}^{(l)}$ in layer $l$. Finally, in order to accomplish the graph convolutional network defined in Equation (\ref{gr_00}), we use the convolutional function $\mathcal{C}$ to convolute the node information as follows
\begin{equation}\label{gr_4}
\bm{\mathrm{X}}^{(l+1)}=\mathcal{C}(\bm{\mathrm{X}}^{(l)})=\bm{\mathrm{X}}^{(l)}\bm{\mathrm{W}}^{(l)},
\end{equation}
where $\bm{\mathrm{W}}^{(l)}$ is the learnable weighting matrix.

\subsection{Graph Transformer}
Sliding windows-based methods \cite{liu2021swin,chu2021twins} model the local relation inside each window via transformer \cite{dosovitskiy2021an}, most of which does not take into account the global modeling of a nonlinear relation between windows. In order to obtain more robust relation-enhanced features, we propose Graph Transformer (GT) to model the relationship between different windows and within each window. We use the visual similarity among different windows to learn the relation of different windows while using the visual similarity between different pixels to model the local relation inside each sliding window. Based on the above-devised efficient graph relation network, the detailed design of the global relation modeling module and the local one will be further elaborated on below. 

\subsubsection{Global Relation Modeling}

We propose a Global Relation (GR) Modeling module to model the global relation among different sliding windows and make the pixel-level classification more accurate. Specifically, we consider each window as one node and use the efficient graph relation network to build the relation of the graph nodes. As a result, we can obtain the coarse relation of the objects in each image via the relation modeling of windows.

Concretely, given the partitioned $M\times N$ sliding windows in feature maps $\bm{\mathrm{X}} \in \mathbb{R}^{C\times H\times W}$, we look at each sliding window $\bm{\mathrm{x}}_{i,j}\in \mathbb{R}^{C\times \frac{H}{M}\times \frac{W}{N}}$ as a node to establish a graph network. For the sake of narrative convenience, we denote $\bm{\mathrm{x}}_{m,n}$ as $\bm{\mathrm{x}}_{i}$, where $i = (m-1)\times N + n$. 

We first define the connection relation matrix of nodes ${\bm{\mathrm{R}}}^{gr}=\{r_{i,j}^{gr}|i,j\in\{1,2,\cdots, M\times N\}\}$, here we use the matrix multiplication-based visual similarity to model the connection relation between nodes. From Equation (\ref{gr_0}), the relation matrix ${\bm{\mathrm{R}}}^{gr}=\mathcal{R}(\bm{\mathrm{X}})$. Simultaneously, in order to alleviate the complexity of visual similarity computation, we devise a function $\mathcal{A}^{gr}$ to reduce the channel dimension
\begin{figure}[t]
\centering
\subfloat[Series]{\includegraphics[width=0.3\linewidth]{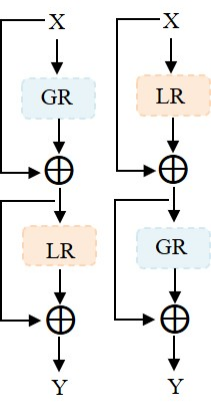}
\label{fig:diff-fusion_a}}
\subfloat[Parallel]{\includegraphics[width=0.3\linewidth]{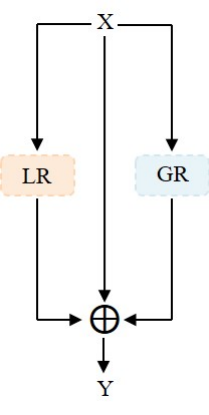}
\label{fig:diff-fusion_b}}

\caption{Three designs of fusion. ''GR'' denotes the global relation modeling module and ''LR'' denotes the local relation modeling module. (a) The two relation modeling modules are connected in two different series connections. (b) Two relation modeling modules are connected in parallel.}
\label{diff-fusion}
\end{figure}

\vspace{-6mm}
\begin{align}
\label{ga_1}
\bm{\mathrm{\hat{x}}}_{i} = {\mathcal{A}^{gr}}(\bm{\mathrm{x}}_{i}),
\end{align}

\noindent where $\bm{\mathrm{\hat{X}}}=\{\bm{\mathrm{\hat{x}}}_{i}\in \mathbb{R}^{\frac{C}{r^{gr}}\times \frac{H}{M}\times \frac{W}{N}}|i\in\{1, 2, \cdots, M\times N\}\}$ is the hidden vector used to compute the connectivity of the graph, mainly to save computation and to enhance the learnability of the module. $r^{gr}$ is the channel compression ratio. Thus, this allows us to derive a formula for calculating the connection relation that saves significant computational complexity, as shown below

\vspace{-6mm}
\begin{align}
\label{ga_0}
\bm{\mathrm{R}}^{gr}=\mathcal{R}^{gr}(\bm{\mathrm{\hat{X}}}),
\end{align}

where, for ease of understanding, we rewrite the relational modeling function $\mathcal{R}$ in Equation ($\ref{gr_0}$) here as $\mathcal{R}^{gr}$. With the connection relation matrix $\bm{\mathrm{R}}^{gr}$ (we rewrite $\bm{\mathrm{R}}$ as $\bm{\mathrm{R}}^{gr}$ in layer $l$ of the graph neural network for simplicity in the following section), based on Equation (\ref{gr_2}) and Equation (\ref{gr_4}), we can define the node update function $\mathcal{U}^{gr}$ and the convolutional function $\mathcal{C}^{gr}$ for global window-aware relation network as follows
\begin{equation}\label{global_2}
\bm{\mathrm{\hat{X}}}^{(l+1)}=\mathcal{C}^{gr}\circ\mathcal{U}^{gr}(\bm{\mathrm{\hat{X}}}^{(l)}),
\end{equation}
where $\circ$ denotes the composition operations on multiple functions, generating a compound function. In order to keep the input and output dimensions unchanged, we define $(\mathcal{A}^{gr})^{-1}$, the inverse transformation of $\mathcal{A}^{gr}$, to restore the dimension of $\bm{\mathrm{\hat{X}}}$ in layer $(l+1)$, as shown below 
\begin{equation}\label{ga_3}
\bm{\mathrm{X}}^{(l+1)}=(\mathcal{A}^{gr})^{-1}(\bm{\mathrm{\hat{X}}}^{(l+1)}).
\end{equation}

\subsubsection{Local Relation Modeling}
In order to learn the local relation inside each sliding window, we conduct a Local Relation (LR) Modeling module to build relations inside each sliding window. Similar to the global relation modeling module, we consider each pixel as a node and exploit the efficient graph relation network to model the local relation among different pixels within each window. 

Specifically, given the sliding window $\bm{\mathrm{x}}_{i,j}\in \mathbb{R}^{C\times \frac{H}{M}\times \frac{W}{N}}$, we need to learn the relationships among these feature points inside each window $\bm{\mathrm{x}}_{i,j}$. Thus, we can similarly define the entire process as follows
\begin{equation}\label{local_0}
\bm{\mathrm{x}}_{i,j}^{(l+1)}=(\mathcal{A}^{lr})^{-1}\circ\mathcal{C}^{lr}\circ\mathcal{U}^{lr}\circ\mathcal{R}^{lr}\circ\mathcal{A}^{lr}(\bm{\mathrm{x}}_{i,j}^{(l)}),
\end{equation}
where $\bm{\mathrm{x}}_{i,j}^{(l)}$ and $\bm{\mathrm{x}}_{i,j}^{(l+1)}$ are the feature maps in layer $l$ and layer $l+1$ of the local window-aware module, respectively.

\subsubsection{Module Structure} As shown in Figure \ref{fig:cib_arch}, our presented graph transformer can be implemented by two $1\times 1$ Conv, a normalization function (Softmax) and other basic elements (Self-Attention and Multilayer Perceptron) in ViT \cite{dosovitskiy2021an}. Specifically, $\mathcal{A}^{gr}$ and $\mathcal{A}^{lr}$ can be completed by a $1\times 1$ Conv, which is used to reduce the computational complexity in the relation modeling process. Similarly, $(\mathcal{A}^{gr})^{-1}$ and $(\mathcal{A}^{lr})^{-1}$ can be realized by another $1\times 1$ Conv, aiming to resume the channel dimension for module integration in the corresponding relation modeling network. In order to accomplish $\mathcal{R}^{gr}$ and $\mathcal{R}^{lr}$, we use a matrix multiplication and a softmax function to norm the results and obtain relation matrix $\bm{\mathrm{R}}^{gr}$ and $\bm{\mathrm{R}}^{lr}$ respectively. The node update function $\mathcal{U}^{gr}$ and $\mathcal{U}^{lr}$ can be accomplished by a matrix multiplication simply.
\subsubsection{Fusion}
Given a global relation modeling module and a local one, we fuse these two different relation modeling modules to obtain better performance. As shown in Figure \ref{diff-fusion}, there are three fusion types, namely, global then local series and local then global series connection (Figure \ref{fig:diff-fusion_a}), parallel connection (Figure \ref{fig:diff-fusion_b}). We will discuss in detail the impact of different fusion methods on segmentation performance in the experimental section.

\begin{figure*}[ht]
	\centering
	\includegraphics[width=0.8\linewidth]{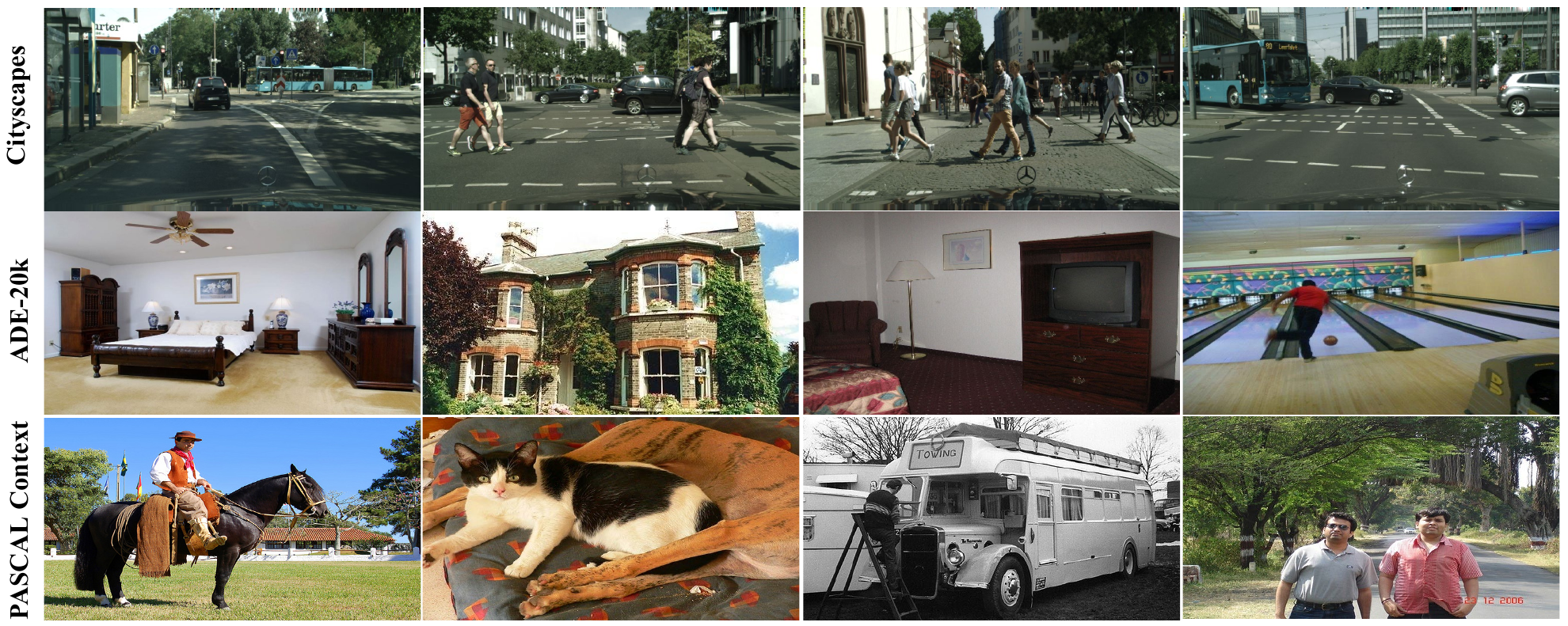}
	\caption{Some examples of the three datasets: Cityscapes (first row), ADE-20k (second row), and PASCAL Context (third row).}
	\label{fig:dataset_show_2}
\end{figure*}

\subsection{Boundary-aware Attention}
A robust feature extraction backbone network has been constructed, as indicated above. We further devise boundary-aware attention (BA) network to optimize the boundary region of the corresponding object for a better overall semantic segmentation effect.

As we all know, the edge information adjustment of objects is generally local, and is greatly affected by local features. Based on this, we design a local-aware attention module, namely the BA module, which further improves the boundary segmentation accuracy of the model by enhancing and weakening the weight information of some \textbf{local feature points} (or pixels, here we do not distinguish between pixels and feature points), some of which are \textbf{the boundaries of objects}.
Concretely, we perceive the relation between local pixels by the local perception module and perform local relation modeling, based on which we learn to obtain the attention coefficients of each surrounding pixel. Finally, we use the attention coefficients to weigh the pixels, strengthen the key pixels, and weaken the inessential ones to achieve accurate boundary segmentation. 

Given the input $\bm{\mathrm{Y}} \in \mathbb{R}^{\widetilde{C}\times \widetilde{H}\times \widetilde{W}}$, where $\widetilde{C}$, $\widetilde{H}$ and $\widetilde{W}$ denotes the dimension of channel, height and width respectively, we use the channel squeezing function $\bm{\mathcal{\widetilde{A}}}$ to reduce the dimension of input features, and then use the local modeling function $\bm{\mathcal{G}}$ to learning the local relation, where the pixels around the boundary of the object is adjusted. Besides, we add non-linear function $\bm{\mathcal{H}}$ inside the boundary-aware module to learn the more robust relation and use the inverse function $(\bm{\mathcal{\widetilde{A}}})^{-1}$ to resume the input channel dimension. At last, we normalize the learned attention coefficient via the normalization function $\bm{\mathcal{N}}$, so the whole formulation can be:

\begin{equation}\label{BA_0}
\bm{\mathrm{Z}}=\bm{\mathcal{N}}\circ (\bm{\mathcal{\widetilde{A}}})^{-1}\circ\bm{\mathcal{H}}\circ \bm{\mathcal{G}}\circ \bm{\mathcal{\widetilde{A}}}(\bm{\mathrm{Y}}).
\end{equation}
 
\subsubsection{Module Structure} As shown in Figure \ref{fig:cib_arch}, boundary-aware attention module compose of two layers of $1\times 1$ Conv, a layer of GELU and a layer of Sigmoid. Similar to the design of the graph transformer module, we use $1\times 1$ Conv to accomplish the channel squeezing and resume functions $\bm{\mathcal{\widetilde{A}}}$ and $(\bm{\mathcal{\widetilde{A}}})^{-1}$. Here we simply use $7\times 7$ Conv to implement the local relation modeling function $\bm{\mathcal{G}}$. Finally, function $\bm{\mathcal{H}}$ and $\bm{\mathcal{N}}$ can be completed through the common non-linear layer (GELU) and normalization layer (Sigmoid), respectively.

\doublerulesep 0.1pt
\begin{table}[!tp]
\begin{footnotesize}
	\begin{center}
	
		\caption{The complexity comparison of our Graph-Segmenter model and the original Swin \cite{liu2021swin} model with Swin-L as a backbone in terms of the number of model parameters (\#param) and the amount of multiplication computation (GMac). $^*$ refers to reproduced results by us.}\label{tab:complexity-ana}
		{
		\begin{tabular}{ l c c  p{4cm} }
			\toprule
			{\upshape Method}                              & {\upshape \# param.}   & {\upshape GMac}     \\
			\midrule
			{\upshape Swin$^*$ \cite{liu2021swin}}        & {\upshape 233.66}     & 191.45          \\
			\midrule
			{\upshape Graph-Segmenter (Ours)}            & {\upshape 283.46}         & 195.63 \\
			\bottomrule
	\end{tabular}
	}
	\end{center}
	\vspace{-0.2in}
\end{footnotesize}
\end{table}

\subsection{Discussion}
We discuss the complexity of the proposed model in this section. The analysis of our designed lightweight pluggable Graph-Segmenter, consisting of a Graph Transformer and Boundary-aware Attention, is shown in Table \ref{tab:complexity-ana}. From the table, we can conclude that the increase of our proposed Graph-Segmenter model with the original Swin \cite{liu2021swin} model in terms of the number of model parameters (\#param) and the multiplicative computation is very small. In particular, in terms of multiplication computation, our model only increases the computation of 2.18\% compared to the original model.

\section{Experiments}
\noindent In this section, we evaluate our proposed Graph-Segmenter on three standard semantic segmentation datasets: Cityscapes \cite{cityscapes}, ADE-20k \cite{ade20k} and PASCAL Context \cite{PASCALC}.

\subsection{Datasets and metrics}

\textbf{Cityscapes} \cite{cityscapes}. The dataset covers a range of 19 semantic classes from 50 different cities. It has 5,000 images in total, with 2,975 for training, 500 for validation, and another 1,525 for testing.

\textbf{ADE-20k} \cite{ade20k}. The dataset annotates 150 categories in challenging scenes. It contains more than 25,000 finely annotated images, split into 20,210, 2,000 and 3,352 for training, validation and testing respectively. 

\textbf{PASCAL Context} \cite{PASCALC}. The dataset is an extension of the PASCAL VOC 2010, which includes 4998 and 5105 photos for training and validation, respectively, and gives pixel-by-pixel semantic labels. It contains more than 400 classes from three major categories (stuff, hybrids and objects). However, since most classes are too sparse, we only analyze the most common 60 classes, which is the approach previous works did \cite{zheng2021rethinking}.

Figure \ref{fig:dataset_show_2} shows some examples of the three datasets.

\textbf{Metrics}. We adopt mean Intersection over Union (mIoU) to evaluate semantic segmentation models.

\subsection{Implementation Details}
The entire model is implemented based on the Swin \cite{liu2021swin} codes that have been made publicly available. Training is separated into two phases: we first establish the backbone network using the model parameters pre-trained on ImageNet, and then we train the backbone network with the Graph-Segmenter that we proposed. The trained network parameters are utilized to initialize and train the network integrated with the segmentation boundary optimization module, which is then used to optimize the segmentation boundary.

\subsection{Comparison with State-of-the-Art}

We compare our proposed Graph-Segmenter with the state-of-the-art models, which contains CNN (e.g., ResNet-101, ResNeSt-200) based methods \cite{chen2018encoder,fu2019dual,yuan2020segmentation,yin2020disentangled}, and latest transformer based methods \cite{liu2021swin,zheng2021rethinking,strudel2021segmenter} on Cityscapes, ADE-20k and PASCAL Context datasets.

\doublerulesep 0.1pt
\begin{table}[h]
\begin{footnotesize}
	\begin{center}
		
		\caption{Multi-scale inference results of semantic segmentation on the Cityscapes validation dataset compared with state-of-the-art methods. $^*$ refers to reproduced results.}
		\label{tab:perf-cityscape-val}
		
		{   
				\begin{tabular}{ l l  c  }
					\toprule 
					
					{\upshape Method}  & {\upshape Backbone}   & {\upshape val mIoU}   \\
					\midrule
					{\upshape FCN \cite{fcn}}  & {\upshape ResNet-101}  & 76.6         \\
	                {\upshape Non-local \cite{wang2018non}}  & {\upshape ResNet-101}  & 79.1         \\
					
					{\upshape DLab.v3+ \cite{chen2018encoder}}  & {\upshape ResNet-101}      & 79.3          \\
					{\upshape DNL \cite{yin2020disentangled}}  & {\upshape ResNet-101}      & 80.5          \\
					{\upshape DenseASPP \cite{yang2018denseaspp}}  & {\upshape DenseNet}  & 80.6  \\
					{\upshape DPC \cite{chen2018searching}}          & {\upshape Xception-71}  & 80.8   \\
					{\upshape CCNet \cite{huang2019ccnet}}   & {\upshape ResNet-101}    & 81.3   \\
					{\upshape DANet \cite{fu2019dual}}       & {\upshape ResNet-101}  & 81.5    \\
					{\upshape Panoptic-Deeplab \cite{cheng2020panoptic}}   & {\upshape Xception-71}    & 81.5   \\
					
					{\upshape Strip Pooling \cite{hou2020strip}}   & {\upshape ResNet-101}   & 81.9   \\
					
					\midrule
					{\upshape Seg-L-Mask/16 \cite{strudel2021segmenter}} & {\upshape ViT-L} &81.3 \\
					{\upshape SETR-PUP \cite{zheng2021rethinking}} & {\upshape ViT-L}  & 82.2 \\
					
					
					{\upshape Swin$^*$ \cite{liu2021swin}}            & {\upshape Swin-L}        &  82.3 \\
					\midrule
					{\upshape Graph-Segmenter (Ours)}            &  {\upshape Swin-L}        &  \textbf{82.9} \\
					\bottomrule
			\end{tabular}
		}
	\end{center}
\end{footnotesize}
\end{table}

\doublerulesep 0.1pt
\begin{table}[ht]
\begin{footnotesize}
	\begin{center}
		\caption{Results of semantic segmentation on the Cityscapes test dataset compared with state-of-the-art methods. $^*$ refers to reproduced results by us.}\label{tab:perf-cityscape}
		{   
			\begin{tabular}{ l  l  c  }
				\toprule
				{\upshape Method}  & {\upshape Backbone}  & {\upshape test mIoU}     \\
				\midrule
				{\upshape PSPNet \cite{zhao2017pyramid}}          & {\upshape ResNet-101}   & 78.4  \\
				{\upshape BiSeNet \cite{yu2018bisenet}}          & {\upshape ResNet-101}   & 78.9          \\
				{\upshape PSANet \cite{zhao2018psanet}}          & {\upshape ResNet-101}   & 80.1         \\
				{\upshape OCNet \cite{yuan2018ocnet}}  & {\upshape ResNet-101}  & 80.1  \\
                {\upshape BFP \cite{ding2019boundary}}      & {\upshape ResNet-101}       & 81.4       \\
				{\upshape DANet \cite{fu2019dual}}      & {\upshape ResNet-101}       & 81.5       \\
				 {\upshape CCNet \cite{huang2019ccnet}}  & {\upshape ResNet-101}  & 81.9  \\
				\midrule
				{\upshape SETR-PUP \cite{zheng2021rethinking}} & {\upshape ViT-L}  & 81.1\\
				{\upshape Swin$^*$ \cite{liu2021swin}}            & {\upshape Swin-L}        &  80.6 \\
				\midrule
				{\upshape Graph-Segmenter (Ours) }            & {\upshape Swin-L}        & \textbf{81.9} \\
				\bottomrule
		    \end{tabular}
		}
	\end{center}
\end{footnotesize}
\end{table}


\doublerulesep 0.1pt
\begin{table}[t]
\begin{footnotesize}
	\begin{center}
		\caption{Multi-scale inference results of semantic segmentation on the ADE-20k validation dataset compared with state-of-the-art methods. $^*$ refers to reproduced results by us.}\label{tab:ade_val}
		{
		\begin{tabular}{ l l   c c c}
		\toprule
			{\upshape Method}                              & {\upshape Backbone}      & {val mIoU}  \\\midrule
			{\upshape FCN \cite{fcn}}   & {\upshape ResNet-101}   & 41.4 \\
			{\upshape UperNet \cite{xiao2018unified}}   & {\upshape ResNet-101}   & 44.9  \\
			{\upshape DANet \cite{fu2019dual}}          & {\upshape ResNet-101}   & 45.3     \\
			{\upshape OCRNet \cite{yuan2020segmentation}}            & {\upshape ResNet-101} & 45.7   \\
			{\upshape ACNet \cite{fu2019adaptive}}          & {\upshape ResNet-101}   & 45.9  \\
			{\upshape DNL \cite{yin2020disentangled}}       & {\upshape ResNet-101}   & 46.0  \\

			{\upshape DLab.v3+ \cite{chen2018encoder}}      & {\upshape ResNeSt-101}      & 47.3        \\
			{\upshape DLab.v3+ \cite{chen2018encoder}} & {\upshape ResNeSt-200} & 48.4  \\
			\midrule
			{\upshape SETR-PUP \cite{zheng2021rethinking}} & {\upshape ViT-L}   & 50.3 \\
			{\upshape SegFormer-B5 \cite{xie2021segformer}} & {\upshape SegFormer}   & 51.8 \\
			{\upshape Seg-L-Mask/16 \cite{strudel2021segmenter}} & {\upshape ViT-L}   & 53.6 \\
			{\upshape Swin$^*$ \cite{liu2021swin}} & {\upshape Swin-L}   & 53.1 \\

			\midrule
			{\upshape Graph-Segmenter (Ours)}            & {\upshape Swin-L}         & \textbf{53.9} \\
			\bottomrule
	\end{tabular}
	}
	\end{center}
\end{footnotesize}
\end{table}

\doublerulesep 0.1pt
\begin{table}[ht]
\begin{footnotesize}
	\begin{center}
	
		\caption{Results of semantic segmentation on the ADE-20k test dataset compared with state-of-the-art methods. $^*$ refers to reproduced results by us.}\label{tab:ade_test}
		{
		\begin{tabular}{ l l  c  p{4cm} }
			\toprule
			{\upshape Method}                              & {\upshape Backbone}   & {\upshape test score}     \\
			\midrule
			
			{\upshape ACNet \cite{fu2019adaptive}}          & {\upshape ResNet-101}    & 38.5  \\
			{\upshape DLab.v3+ \cite{chen2018encoder}} & {\upshape ResNeSt-101}  & 55.1 \\
			{\upshape OCRNet \cite{yuan2020segmentation}}   & {\upshape ResNet-101}    & 56.0  \\
			{\upshape DNL \cite{yin2020disentangled}}       & {\upshape ResNet-101}    & 56.2   \\
			\midrule
			{\upshape SETR-PUP \cite{zheng2021rethinking}} & {\upshape ViT-L}    & 61.7\\
			{\upshape Swin$^*$ \cite{liu2021swin}}        & {\upshape Swin-L}     & 61.5            \\
			\midrule
			{\upshape Graph-Segmenter (Ours)}            & {\upshape Swin-L}         & \textbf{62.4} \\
			\bottomrule
	\end{tabular}
	}
	\end{center}
\end{footnotesize}
\end{table}

\begin{figure*}[t]
	\centering
	\includegraphics[width=0.85\linewidth]{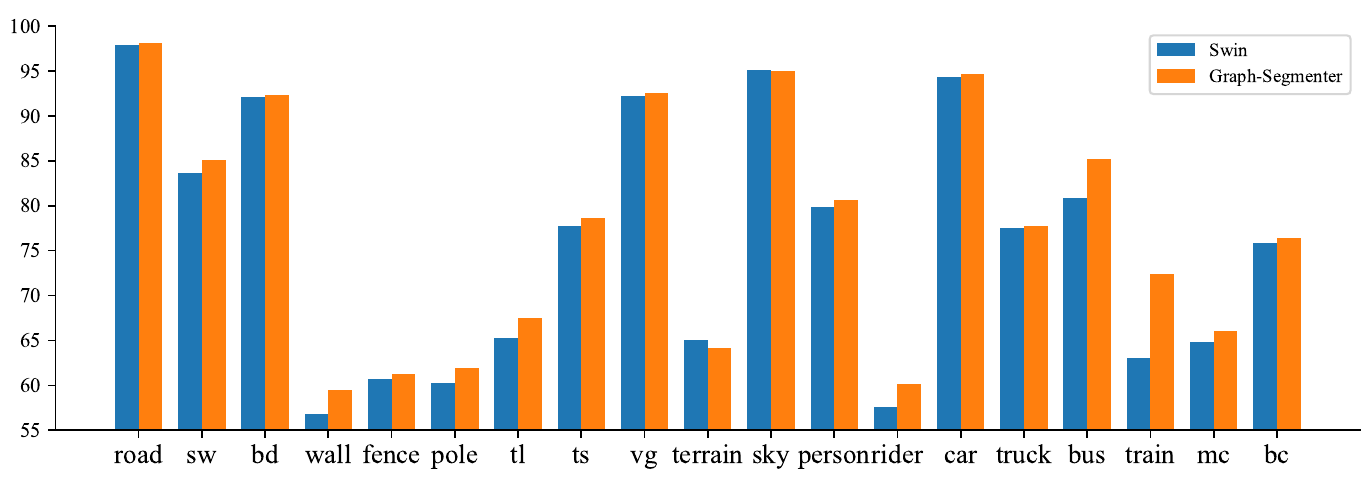}
	\caption{Val mIoU comparison with Swin Tranformer \cite{liu2021swin} in each class on the Cityscapes val set. Our Graph-Segmenter generally achieves better performance. "sw" denotes "sidewalk", "bd" denotes "building", "tl" denotes  "traffic light", "ts" denotes "traffic sign", "vg" denotes  "vegetation", "mc" denotes "motorcycle" and "bc" denotes "bicycle".}
	\label{fig:class-compare}
\end{figure*}

\textbf{Cityscapes}. Table \ref{tab:perf-cityscape-val} and Table \ref{tab:perf-cityscape} demonstrate the results on Cityscapes dataset. It can be seen that our Graph-Segmenter is +0.6 mIoU higher (82.9 vs. 82.3) than transformer-based method Swin-L \cite{liu2021swin} in Cityscapes validation dataset and +0.8 mIoU higher (81.9 vs. 81.1) than transformer-based SETR-PUP \cite{zheng2021rethinking} in Cityscapes test dataset. In addition, we further show the accuracy of our proposed Graph-Segmenter in each category, as shown in Figure \ref{fig:class-compare}. From the figure, we can see that Graph-Segmenter almost obtains the best performance in all categories compared to the Swin model, and even for some more difficult categories (e.g., train, rider, wall) our model still achieves a very large performance improvement. This phenomenon shows that global and local window relation modeling and boundary-aware attention modeling have further improved the ability of the network for difficult-to-segment categories.

\begin{figure*}[t]
	\centering
	\includegraphics[width=15cm,height=6.2cm]{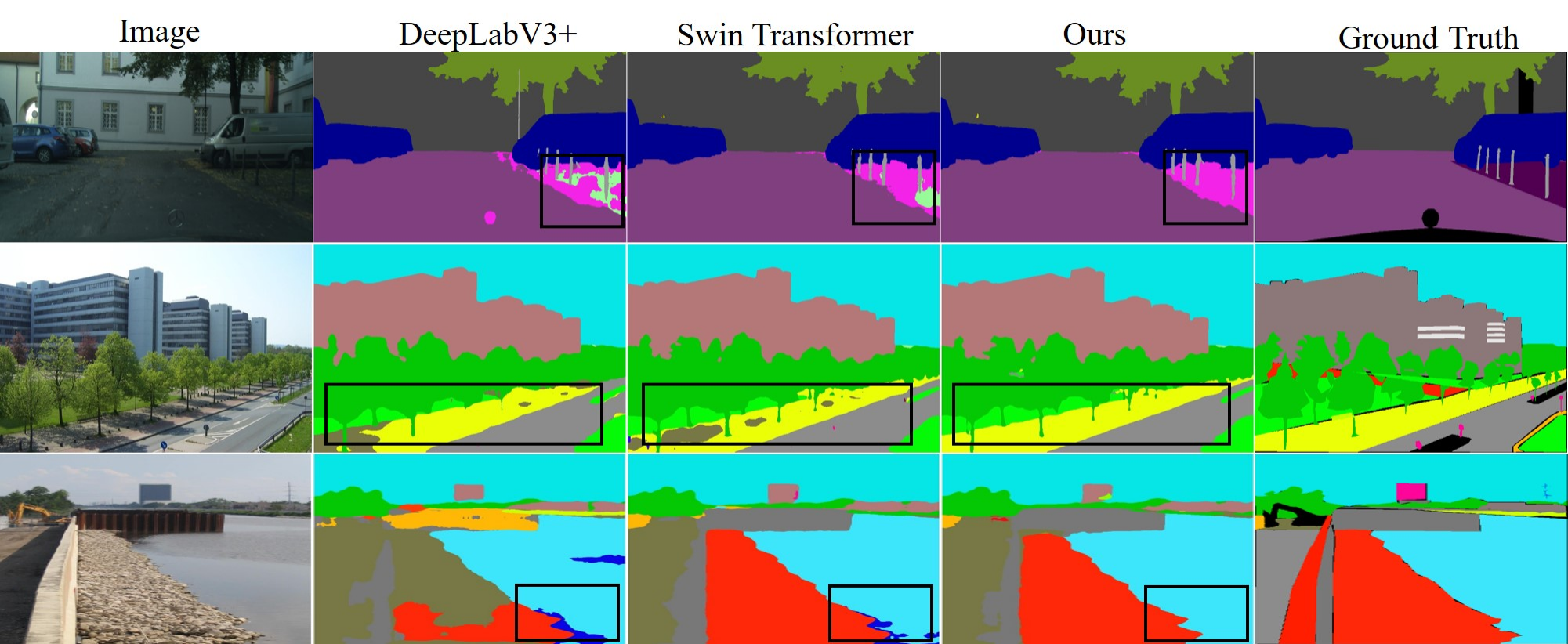}
	\vspace{-0in}
	\caption{Qualitative performance comparison of our proposed Graph-Segmenter with DeepLabV3+ \cite{chen2018encoder} and Swin Tranformer \cite{liu2021swin} for semantic segmentation. Our Graph-Segmenter can obtain a better segmentation boundary.}
	\label{fig:boundary-compare}
\end{figure*}

\textbf{ADE-20k}. Table \ref{tab:ade_val} and Table \ref{tab:ade_test} demonstrate the results on ADE-20k dataset.
Our Graph-Segmenter achieves 53.9 mIoU and 62.4 test score on the ADE-20k val and test set, surpassing the previous best transformer-based model by +0.3 mIoU (53.6 mIoU by Seg-L-Mask/16 \cite{strudel2021segmenter}) and +0.7 test score (61.7 by SETR-PUP \cite{zheng2021rethinking}). Our approach still can perform excellently in the challenging ADE-20k dataset with more categories, which confirms our modules' robustness. A more qualitative comparison is shown in Figure
\ref{fig:boundary-compare}.


\textbf{PASCAL Context}. To further demonstrate the effectiveness of our proposed Graph-Segmenter module, we conduct more experimental results on PASCAL Context dataset for UperNet \cite{upernet2018} and ``UperNet + CAR'' \cite{huang2022car} models. As shown in Table \ref{tab:pascal_test}, we can observe that different models equipped with our Graph-Segmenter module can gain a consistent performance improvement. All in all, the very recent model ``UperNet + CAR'' embedding Graph-Segmenter achieves the best mIoU accuracy.


\doublerulesep 0.1pt
\begin{table}[h]
\begin{footnotesize}
	\begin{center}

		\caption{Results of semantic segmentation on the Pascal Context dataset compared with state-of-the-art methods.}\label{tab:pascal_test}
		{
		
		\begin{tabular}{ l l  c  p{4cm} }
			\toprule
			{\upshape Method}                              & {\upshape Backbone}   & {\upshape mIoU}     \\
			\midrule

			{\upshape FCN\cite{fcn}} & {\upshape ResNet-101}  & 45.74 \\
			{\upshape PSPNet \cite{zhao2017pyramid}}   & {\upshape ResNet-101}    & 47.80  \\
			{\upshape DANet \cite{fu2019dual}}       & {\upshape ResNet-101}    & 52.60   \\
			{\upshape EMANet \cite{li2019expectation}} & {\upshape ResNet-101}    & 53.10\\
			{\upshape SVCNet \cite{ding2019semantic}}        & {\upshape ResNet-101}     & 53.20            \\
            {\upshape BFP \cite{ding2019boundary}}      & {\upshape ResNet-101}       & 53.6       \\
			{\upshape Strip pooling \cite{hou2020strip}}        & {\upshape ResNet-101}     & 54.50   \\
			{\upshape GFFNet \cite{li1904gff}}        & {\upshape ResNet-101}     & 54.20            \\
			{\upshape APCNet \cite{he2019adaptive}}        & {\upshape ResNet-101}     & 54.70            \\
			{\upshape GRAr \cite{ding2021interaction}}        & {\upshape ResNet-101}     & 55.70            \\
			
			\midrule
			{\upshape SETR-PUP\cite{zheng2021rethinking}}        & {\upshape ViT-L}     & 55.83 \\ 
			\midrule

			{\upshape  UperNet \cite{upernet2018}}        & {\upshape Swin-L}     & 57.48 \\
			
			\midrule
			{\upshape Graph-Segmenter + UperNet (Ours)}            & {\upshape Swin-L}         & \textbf{57.80} \\
			{\upshape Graph-Segmenter + UperNet + CAR (Ours)}            & {\upshape Swin-L}         & \textbf{59.01} \\
			
			\bottomrule
	\end{tabular}
	
		}

	\end{center}
\end{footnotesize}
\end{table}


\textbf{Qualitative results}. We further visualized and analyzed our devised Graph-segmenter network, as shown in Figure \ref{fig:boundary-compare}. As shown in Figure \ref{fig:boundary-compare}, our Graph-Segmenter can obtain a better segmentation boundary compared with DeepLabV3+ \cite{chen2018encoder} and Swin Tranformer \cite{liu2021swin}.

\doublerulesep 0.1pt
\begin{table}[t]
\begin{footnotesize}
	\begin{center}
		\caption{mIoU performance of different connection types on Cityscapes dataset. ``$\cup$'' denotes parallel connection. ``-T'' denotes Tiny.}\label{tab:perf-diff-connect-type}
		{   
			\begin{tabular}{ l l  c  }
			    \toprule
				{\upshape Connection Type} & {\upshape Backbone}        & {\upshape val mIoU}  \\\midrule
				{\upshape GR $\cup$ LR}     & {\upshape Swin-T}    &  76.70      \\
				{\upshape LR followed by GR}   & {\upshape Swin-T}      &  76.44             \\
				{\upshape GR followed by LR}    & {\upshape Swin-T}     &  \textbf{76.99}            \\
				\bottomrule
		    \end{tabular}
		    }
	\end{center}
\end{footnotesize}
\end{table}

\doublerulesep 0.1pt
\begin{table}[t]
\begin{footnotesize}
\renewcommand\arraystretch{1.3}

	\begin{center}
		
		\caption{mIoU performance of different $\theta$ on Cityscape dataset. We set $v = \frac{1}{K^2}\displaystyle\sum_{i}^{K}\sum_{j}^{K} r_{i,j}$ for the convenience of expression. ``-T'' denotes Tiny.}\label{tab:perf-sparse-ana}
		{   
			\begin{tabular}{ l  l  c   }
			
				\toprule
				{\upshape $\theta$}                              & {\upshape Backbone}      & {\upshape val mIoU}      \\\midrule
				{\upshape $2v$}        & {\upshape Swin-T}    &  76.44  \\
				{\upshape $v$}        & {\upshape Swin-T}    &   76.27    \\
				{\upshape $\frac{1}{2}v$}        & {\upshape Swin-T}    & 76.07 \\
				{\upshape $\frac{1}{4}v$}        & {\upshape Swin-T}    &  77.64 \\
				{\upshape $\frac{1}{8}v$}     & {\upshape Swin-T}        &  77.05   \\
				\bottomrule
		    \end{tabular}
		}
	\end{center}

	\vspace{-0.2in}
\end{footnotesize}
\end{table}
\doublerulesep 0.1pt
\begin{table}[b]
\begin{footnotesize}
	\begin{center}
		
		\caption{mIoU performance of different components on Cityscape dataset. ``-T'' denotes Tiny.}\label{tab:perf-diff-compponent}
		{   
			\begin{tabular}{ l  l  c   }
			
				\toprule
				{\upshape Method}                              & {\upshape Backbone}      & {\upshape val mIoU}      \\\midrule
				{\upshape Swin \cite{liu2021swin}}        & {\upshape Swin-T}    &  75.82     \\
				{\upshape Swin-GT}        & {\upshape Swin-T}    &  76.99    \\
				{\upshape Swin-BA}        & {\upshape Swin-T}    &  75.94 \\
				{\upshape Swin-GTBA }     & {\upshape Swin-T}        &  \textbf{77.32}   \\
				\bottomrule
		    \end{tabular}
		}
	\end{center}
\end{footnotesize}
\end{table}

\subsection{Ablation Study}
In this section, we deeply investigate the performance of each sub-module of the model and conduct experiments on the Cityscapes dataset.

\subsubsection{Sparsity Analysis} We first analyze the effect of different sparsity of the relation matrix $\bm{\mathrm{R}}$ on the experimental results. In order to eliminate the influence of other factors, we do not add the boundary-aware attention module in our experiments. By a simple setting of $\theta = \frac{1}{K^2}\displaystyle\sum_{i}^{K}\sum_{j}^{K} r_{i,j}$, the accuracy of the model with the sparse constraint on the Cityscapes dataset could achieve 0.59 rise than the case without it. It can be concluded that for a given node, not all surrounding nodes can provide positive feature enhancement for the node. On the contrary, by properly filtering out some nodes, the model can achieve better segmentation performance.
 To further discuss the effect of the value of $\theta$ on the experimental results, we set $\theta$ as different thresholds, respectively, as shown in Table \ref{tab:perf-sparse-ana}. From the table, we can observe that different thresholds still have a certain degree of influence on the final segmentation effect of the model, and it is found that the model can achieve the best results when $\theta = \frac{1}{4K^2}\displaystyle\sum_{i}^{K}\sum_{j}^{K} r_{i,j}$. Besides, the experimental results further prove the effectiveness of the sparse setting, which can not only simplify the calculation, but also improve the segmentation accuracy of the model to a certain extent.

\subsubsection{Different Connection Type Analysis} We analyze the influence of the different sparseness of the relation matrix R on the experimental results. As Table \ref{tab:perf-diff-connect-type} shows, we can see that the connection in the serial is better than in parallel. Meanwhile, the best performance is obtained for serial connections by placing global window-aware attention before local window-aware attention. The phenomenon further indicates that better segmentation results can be achieved by large-scale coarse-grained relation modeling among windows first and following fine-grained modeling within each window.

\doublerulesep 0.1pt
\begin{table}[b]
\begin{footnotesize}
	\begin{center}
		\caption{mIoU performance of two components in the graph transformer module on Cityscape dataset. ``-T'' denotes Tiny.
		}\label{tab:perf-diff-GT}
		{   
			\begin{tabular}{ l  l  c   }
				\toprule
				{\upshape Method}                              & {\upshape Backbone}      & {\upshape val mIoU}      \\\midrule
				{\upshape Swin \cite{liu2021swin}}        & {\upshape Swin-T}    &  75.82     \\
				{\upshape Swin-GR}        & {\upshape Swin-T}    &  76.24    \\
				{\upshape Swin-LR}        & {\upshape Swin-T}    &  76.28\\
				{\upshape Swin-GT}     & {\upshape Swin-T}        &  \textbf{76.99}   \\
				\bottomrule
		    \end{tabular}
		}
	\end{center}
\end{footnotesize}
\end{table}

\doublerulesep 0.1pt
\begin{table}[b]
\begin{footnotesize}
	\begin{center}
		\caption{mIoU performance of Graph-Segmenter (only with graph transformer module) on Cityscapes dataset for various channel compression ratio $r$.}\label{tab:perf-whether-spase}
		{   
			\begin{tabular}{ c  l l l l l  }
			    \toprule
			    r & \, r=2 & \, r=4 & \, r=8 &  r=16 &  r=32\\
				\midrule
				mIoU & 76.65 & 76.15 & 76.29 & 76.99 & 76.35\\
				\bottomrule
		    \end{tabular}
		}
	\end{center}
\end{footnotesize}
\end{table}

\subsubsection{Graph Transformer and Boundary-aware Attention Modules Analysis} In order to analyze the specific role of each component, we compare various sub-modules to analyze the experimental results separately. As Table \ref{tab:perf-diff-compponent} shows, the model equipped with a graph transformer (Swin-GT) module or equipped with boundary-aware attention (Swin-BA) module can obtain a noticeable improvement compared to the original Swin model. Besides, we can observe that BA can improve a lot over Swin-GT, although it only gains marginal improvement over Swin. This phenomenon implies that after modeling the global and local relationship, the BA module can achieve more accurate classification based on more robust features than without Swin-GT enhancement. At the same time, we can see that Swin embeds all of the proposed modules (GT+BA) and obtains the best performance compared to the model embedding only some of our modules.

For analyzing the effect of two components in the graph transformer module, we compare the sub-module results, as shown in Table \ref{tab:perf-diff-GT}. The model with the Global Relation Modeling module (Swin-GR) or with the Local Relation Modeling module (Swin-LR) can obtain a visible improvement compared to the original model. When combining both kinds of sub-modules (Swin-GT), the model can achieve the best performance.

\subsubsection{Different Channel Compression Ratio $r$ Analysis}
We further analyze the performance of different compression ratios in the graph transformer module. As indicated in Table \ref{tab:perf-whether-spase}, we can see that the model obtains the best semantic segmentation accuracy when the compression ratio $r$ (here we simply set $r^{gr}$ and $r^{lr}$ to the same value and mark them as $r$) is 16, which indicates that either too large or too small compression rate will have an impact on the recognition progress of the model, so as shown in the table, we choose the same compression rate of 16 in this paper.

\section{Conclusion}
\noindent This paper presents Graph-Segmenter, which enhanced the vision transformer with hierarchical level graph reasoning and efficient boundary adjustment requiring no additional annotation. Graph-Segmenter achieves state-of-the-art performance on Cityscapes, ADE-20k and PASCAL Context semantic segmentation, noticeably surpassing previous results. Extensive experiments demonstrate the effectiveness of multi-level graph modeling, features with unstructured associations leading to evident rise; and the effectiveness of the Boundary-aware Attention (BA) module, minor refinements around boundary gave smoother masks than the previous state-of-the-art.

\bibliographystyle{unsrt}
\bibliography{egbib}


\Biography{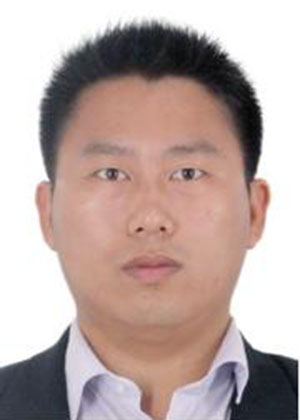}{\textbf{Zizhang Wu} received the B.Sc. and M.Sc. degree in Pattern Recognition and Intelligent Systems from Northeastern University in 2010 and 2012, respectively. He is currently a perception algorithm manager in the Computer Vision Perception Department of ZongMu Technology. He is mainly responsible for the development of core perception algorithms, optimizing algorithms, and model effects, and driving business development with technology.}{width=3cm,height=3.8cm}

\Biography{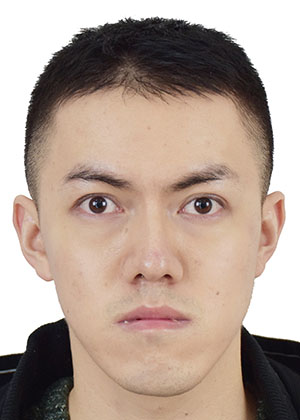}{\textbf{Yuanzhu Gan} received the M.Sc. degree in Pattern Recognition and Artificial Intelligence from Nanjing University, Jiangsu, China, in 2021. He is now an algorithm engineer at ZongMu Technology, Shanghai, China. His current interests include 3D object detection for autonomous driving.}{width=3cm,height=3.8cm}

\Biography{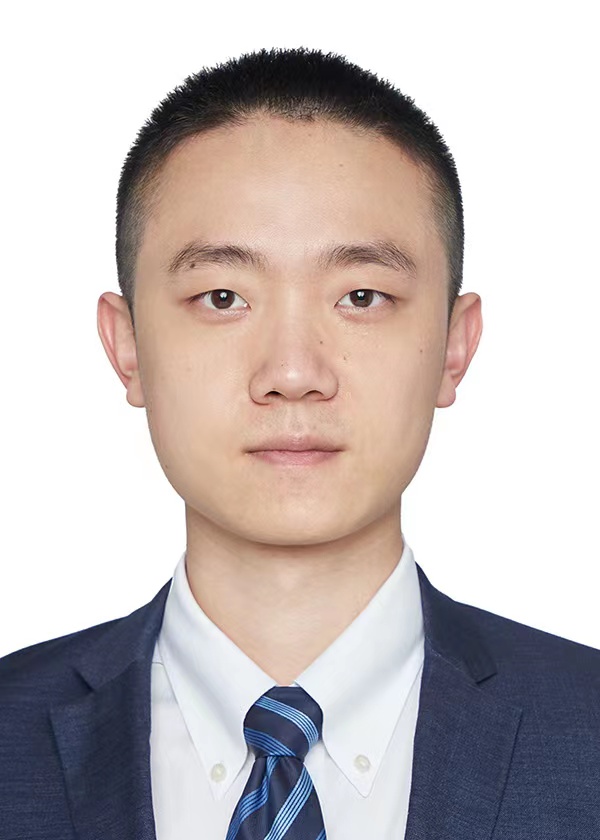}{\textbf{Tianhao Xu} received the B.E. degree in Vehicle Engineering from Jilin University, Jilin, China, in 2017. He is currently pursuing the M.Sc. degree in Electric Mobility at Technical University of Braunschweig, Braunschweig, Germany. His current research interests include computer vision and microstructural analysis in machine learning.}{width=3cm,height=3.8cm}

\Biography{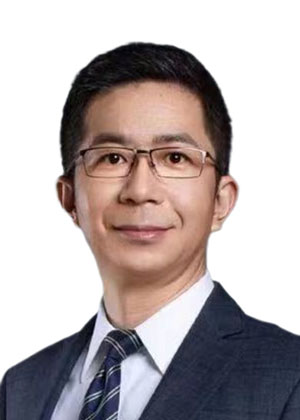}{\textbf{Fan Wang} received the B.Sc. and M.Sc. degree in Computer Science and Artificial Intelligence from Northwestern Polytechnical University, Xi'an, China, in 1997 and 2000, respectively. He is currently with ZongMu Technology as a Vice President and Chief Technology Officer. His current research interests include Computer Vision, Sensor Fusion, Automatic Parking, Planning \& Control, and $L2/L3/L4$ Autonomous Driving.}{width=3cm,height=3.8cm}

\end{document}